\documentclass{article}

\usepackage{spconf,amsmath,graphicx}
\usepackage{multirow}
\usepackage{subfig}
\usepackage{booktabs}
\usepackage{bbding}
\usepackage{amssymb}

\usepackage{pifont}      
\usepackage{bbding}      
\usepackage{fontawesome}  
\usepackage{xcolor}
\usepackage{array}
\usepackage{microtype}
\usepackage[pagebackref=false,breaklinks=true,letterpaper=true,colorlinks,citecolor=blueish,linkcolor=brickred,bookmarks=false]{hyperref}

\definecolor{brickred}{rgb}{0.8, 0.25, 0.33}
\definecolor{blueish}{rgb}{0.0, 0.3, 0.6}

\newcommand{\PreserveBackslash}[1]{\let\temp=\\#1\let\\=\temp}

\newcolumntype{C}[1]{>{\PreserveBackslash\centering}p{#1}}
\newcolumntype{R}[1]{>{\PreserveBackslash\raggedleft}p{#1}}
\newcolumntype{L}[1]{>{\PreserveBackslash\raggedright}p{#1}}

\def\VspaceTc{\vspace{-0.40cm}}

\def\VspacePa{\vspace{-0.30cm}}
\def\VspacePb{\vspace{-0.20cm}}

\newcommand{\cmark}{\ding{51}}

\title{GATOR: Graph-Aware Transformer with Motion-Disentangled Regression for Human Mesh Recovery from a 2D Pose}

\name{Yingxuan You$^1$, Hong Liu$^{1*}$, Xia Li$^2$, Wenhao Li$^1$, Ti Wang$^1$, Runwei Ding$^1$
\thanks{
*Corresponding author: hongliu@pku.edu.cn.
This work is supported by National Natural Science Foundation of China (No.62073004), Basic and Applied Basic Research Foundation of Guangdong (No. 2020A1515110370), Shenzhen Fundamental Research Program (GXWD20201231165807007-20200807164903001, No. JCYJ20200109140410340).}}

\address{
Key Laboratory of Machine Perception, Peking University, Shenzhen Graduate School$^1$ \\
Department of Computer Science, ETH Zürich$^2$ \\
\texttt{\small\{youyx, tiwang\}@stu.pku.edu.cn, \{hongliu, wenhaoli, dingrunwei\}@pku.edu.cn, xia.li@inf.ethz.ch}
}
\begin{document}

\maketitle

\begin{abstract}
3D human mesh recovery from a 2D pose plays an important role in various applications.
However, it is hard for existing methods to simultaneously capture the multiple relations during the evolution from skeleton to mesh, including joint-joint, joint-vertex and vertex-vertex relations, which often leads to implausible results.
To address this issue, we propose a novel solution, called GATOR, that contains an encoder of Graph-Aware Transformer~(GAT) and a decoder with Motion-Disentangled Regression~(MDR) to explore these multiple relations. 
Specifically, GAT combines a GCN and a graph-aware self-attention in parallel to capture physical and hidden joint-joint relations. 
Furthermore, MDR models joint-vertex and vertex-vertex interactions to explore joint and vertex relations. Based on the clustering characteristics of vertex offset fields, MDR regresses the vertices by composing the predicted base motions. 
Extensive experiments show that GATOR achieves state-of-the-art performance on two challenging benchmarks. 
Code is available at \href{https://github.com/kasvii/GATOR}{https://github.com/kasvii/GATOR}.
\end{abstract}

\vspace{-0.10cm}

\vspace{-2.0mm}
\section{Introduction}
\vspace{-2.0mm}

\label{sec:intro}

3D human mesh recovery from the 2D observation is an essential task for many 3D applications~\cite{survey}.
However, image-based methods suffer from the domain gap in image appearance between well-controlled datasets and in-the-wild scenes,
while pose-based methods naturally relieve this problem with the skeleton inputs~\cite{choi2020pose2mesh, pqgcn, zheng2021lightweight, mug}. 
But existing pose-based methods neglect the multiple relations during the evolution from skeleton to mesh, including joint-joint, joint-vertex, and vertex-vertex relations, that are prone to produce implausible results.

Existing pose-based methods follow an encoder-decoder manner~\cite{survey}. In encoders, Graph Convolution Networks~(GCNs) and Transformers have become the mainstream~\cite{zeng2021learning,li2022exploiting,mhformer}. 
Benefiting from the graph structure of human skeleton, GCNs naturally capture the physical relations between neighboring joints~\cite{zou2021modulated,li2022graphmlp}. 
But it is difficult to capture non-local relations.
In contrast, Transformers can
explore global information by the attention mechanism while weakening the graph topology and local relations. 
Recently, several methods combine GCNs and Transformers to complement each other~\cite{zheng2021lightweight, lin2021mesh}.
But the neglect of graph structures in Transformers and the adopted cascaded architecture may limit the effectiveness.

\begin{figure}
  \centering
  \includegraphics[width=0.95\linewidth]{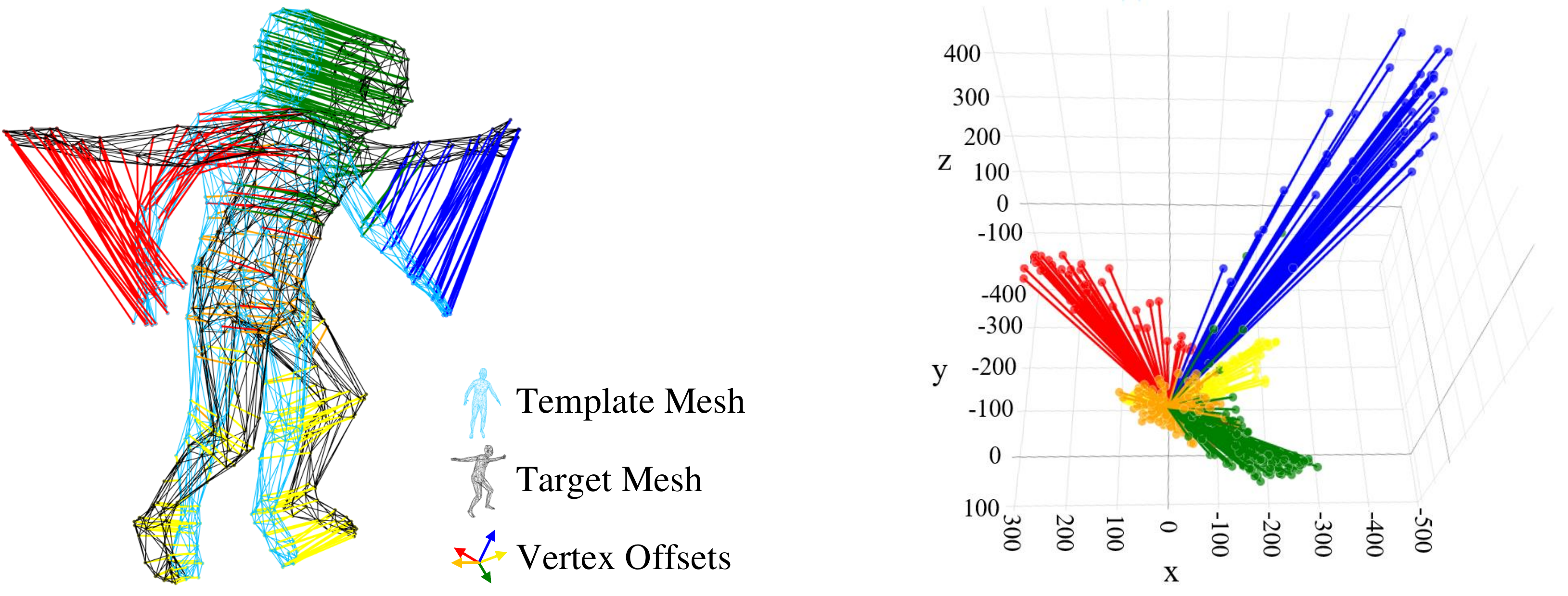}
  \vspace{-4mm}
  \caption{The offset field~(template~$\to$~target)~can be clustered to several base motions drawn in the corresponding colors.
}
  \label{fig_motions}
  \vspace{-4.5mm}
\end{figure}

For decoders, 
some methods regress vertex coordinates~\cite{lin2021mesh, kolotouros2019convolutional,lin2021end}, and some recent works predict the offset fields then add to the template mesh~\cite{zheng2021lightweight, luan2021pc, zanfir2021thundr}. 
They regress 3D coordinates directly from high-dimension features, which is data-driven ignoring the physical plausibility. 
As shown in Fig.~\ref{fig_motions}, the offset field from template mesh to target mesh can be clustered to several base motions due to the motion similarity in the same body part.
This inspires us to generate the vertex motions by predicting the base motions and using them to constitute each vertex offset.
Compared to directly regressing the vertex offsets, predicting and aggregating the base motions shall release the network training burden and provide more accurate results.

\begin{figure*}[htb]
    \centering    
    \includegraphics[width=0.99\linewidth]{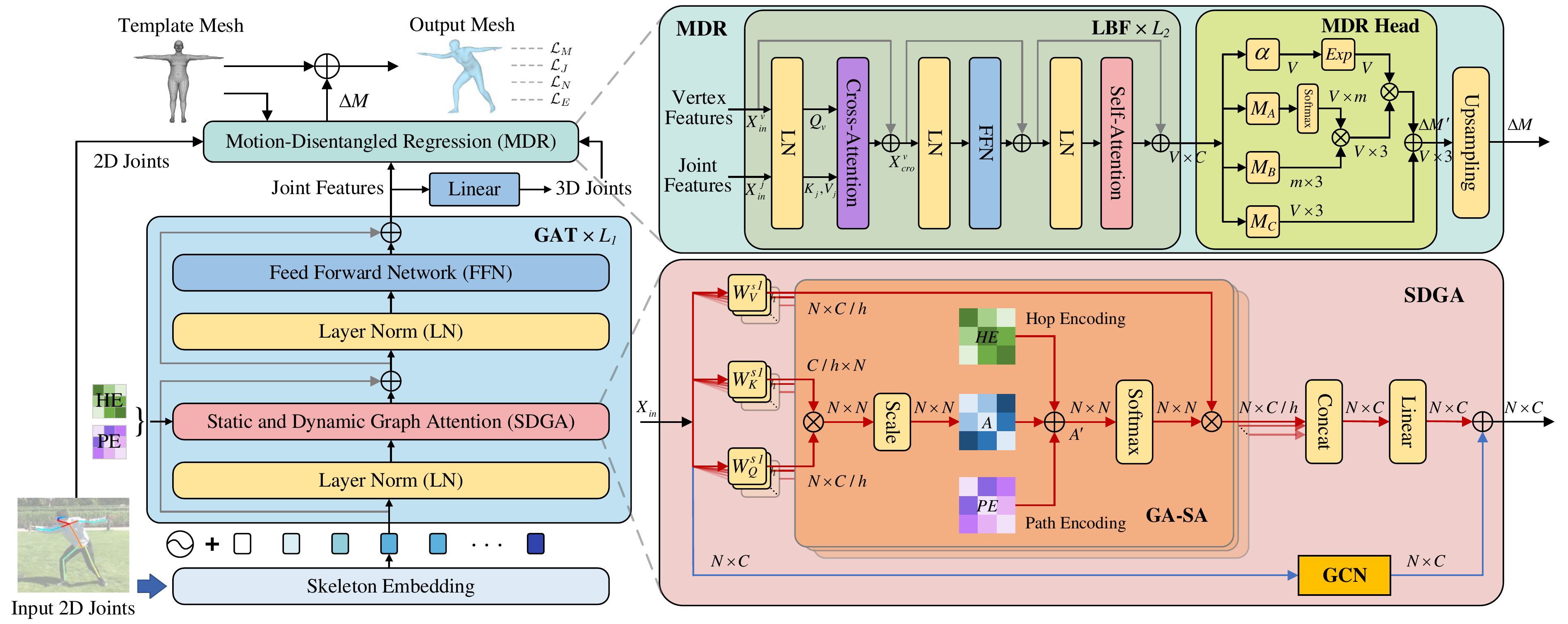}
    \vspace{-3mm}
    \caption{
        Architecture of GATOR.
        Given an input 2D pose,
        the Graph-Aware Transformer~(GAT) encoder learns local-global joint features by the parallel GCN and Graph-Aware Self-Attention~(GA-SA), where GA-SA adds two encodings for skeleton graph awareness. The Motion-Disentangled Regression~(MDR) decoder then generates the vertex offsets by composing base motions.
    }
    \label{fig:overview}
    \vspace{-4mm}
\end{figure*}

Based on the above observations, we present a novel network, termed GATOR, including an encoder of Graph-Aware Transformer (GAT) and a decoder with Motion-Disentangled Regression (MDR), which recovers 3D human mesh from a 2D human pose. (1)~In GAT, we design a two-branch module that contains a GCN branch and a Graph-Aware Self-Attention~(GA-SA) branch to explore physical and hidden joint relations, where GA-SA takes two important skeleton encodings to enhance graph awareness. 
(2)~Moreover, MDR models joint-vertex and vertex-vertex interactions and generates the vertex offsets by composing the predicted base motions.
(3)~Experimental results show that GATOR outperforms previous state-of-the-art methods on two benchmark datasets.

\vspace{-3mm}
\section{Method}
\vspace{-3mm}
\label{sec:method}
Fig.~\ref{fig:overview} illustrates the architecture of GATOR, including GAT and MDR. 
Given 2D human joints estimated by an off-the-shelf 2D pose detector, GAT first extracts physical and hidden joint features and then generates a 3D pose. MDR learns vertex features through joint-vertex and vertex-vertex interactions, then predicts base motions to constitute the vertex offsets which are added to the template mesh as the final mesh.

\VspacePa
\vspace{-1mm}
\subsection{Graph-Aware Transformer Encoder}
\VspacePb
The joint relations include physical skeleton topology and action-specific information (e.g., the relation between hands and feet is strong during running but weak when sitting), which is difficult to capture by a static graph~\cite{zeng2021learning}. Therefore, we propose a two-branch module named Static and Dynamic Graph Attention~(SDGA). One branch is GA-SA which takes two important skeleton structures to improve graph awareness for global and dynamic feature learning. The other is the GCN branch to enhance the physical topology along a static graph.

\noindent \textbf{Graph-Aware Self-Attention.}
Inspired by graph representation tasks~\cite{ying2021transformers}, wherein 
the injected priors in the attention mechanism can adaptively change the attention distribution, we design GA-SA by introducing two crucial skeleton priors. 

One is the multi-hop connectivity between joints, represented by a $N {\times} N$ matrix $D$, where $N$ is the number of joints and $D_{ij} {=} \phi(i,j)$ denotes the hop distance between joint $i$ and joint $j$. 
A learnable embedding table $T_s {\in} \mathbb{R}^{\operatorname{max}(D) \times H}$ is used to project each hop number in $D$ to a vector of size $H$, the head number, and thus embeds the matrix $D$ to a learnable tensor named Hop Encoding~(HE) ${\in} \mathbb{R}^{N \times N \times H}$:
\vspace{-2mm}
\begin{equation}
    \operatorname{HE}_{ij} = T_s[D_{ij}],
    \vspace{-2mm}
\end{equation}
where $[\cdot]$ represents the indexing operation. 

The other is the path information between joints, which reflects the bone length and the body proportion~\cite{survey}. The Path Encoding (PE) mechanism is built upon a distance graph $\{J, E\}$, where $J$ denotes the joints, and $E$ denotes the spatial distances between adjacent joints.
The vector $p_{ij} {=} \{e_{ij}^1, e_{ij}^2, \dots, e_{ij}^{D_{ij}}\}$ is defined as the path from joint $i$ to joint $j$. 
A linear embedding layer $f (\cdot)$ is used to project each path $p_{ij}$ to a learnable tensor: $f(p_{ij}) {\in} \mathbb{R}^{{D_{ij}} \times H}$. The path encoding of joint pair $(i, j)$ is defined as an average of the dot-products of the edge embeddings and the learnable weights in the path:
\vspace{-2mm}
\begin{equation}
    \operatorname{PE}_{ij} = \frac{1}{D_{ij}} \sum_{k=1}^{D_{ij}} W_{ij}^k f(e_{ij}^k),
    \vspace{-2mm}
\end{equation}
where $W_{ij} {\in} \mathbb{R}^{D_{ij}}$ denotes the learnable weights for $p_{ij}$.

By adding up the hop and path encodings to the attention matrix $A$, the improved attention matrix $A'$ can be written as:
\vspace{-1mm}
\begin{equation}
    A'_{ij} = A_{ij} + HE_{ij} + PE_{ij},
\end{equation}
\vspace{-5mm}
\begin{equation}
    A_{ij} = {(X_iW_Q^s)  (X_jW_K^s)^{T} } / {\sqrt{d}},
    \label{SMHA}
    \vspace{-1mm}
\end{equation}
where $d$ is the feature dimension, $X_i, X_j {\in} \mathbb{R}^{d}$ are the input features, and $W_Q^s, W_K^s {\in} \mathbb{R}^{d \times d}$ are the learnable weight matrices that project the input to different representations. 

\begin{figure*}[!t]
  \includegraphics[width=\linewidth]{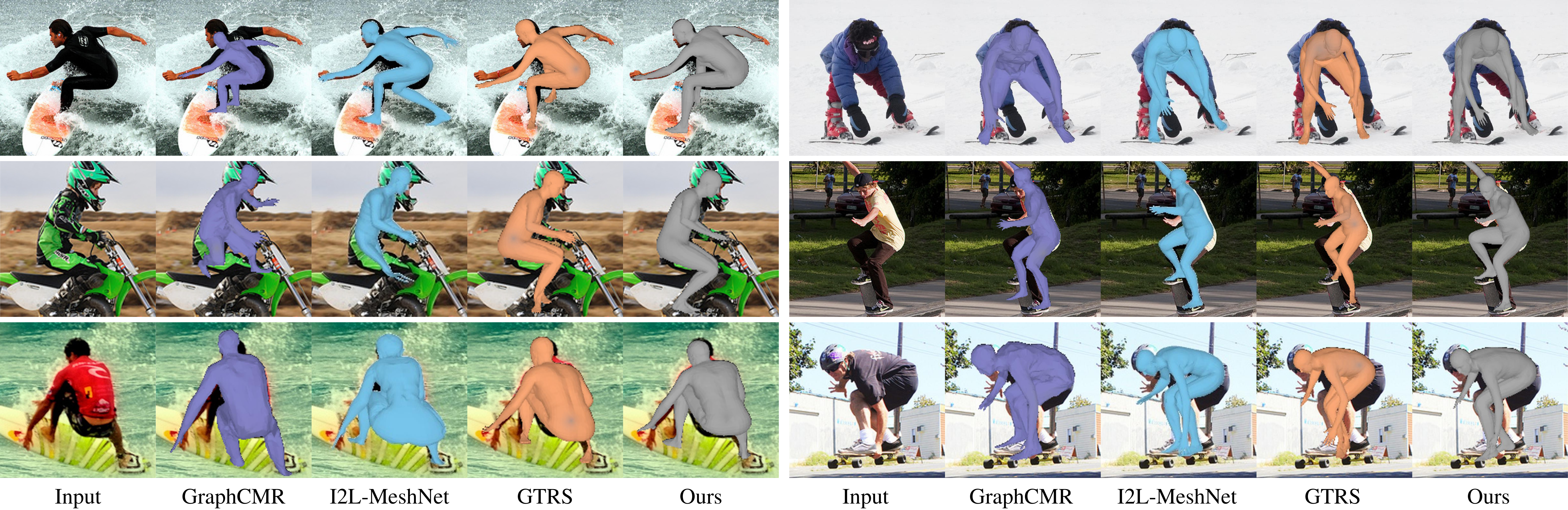}
  \vspace{-6mm}
  \caption{Qualitative results on COCO. From left to right: input image, GraphCMR~\cite{kolotouros2019convolutional}, I2L-MeshNet~\cite{moon2020i2l}, GTRS~\cite{zheng2021lightweight}, and ours.}
  \label{fig_vis}
  \vspace{-4mm}
\end{figure*}

\noindent \textbf{Static and Dynamic Graph Learning.}
GA-SA is updated by input features in both training and inference processes. When the input pose varies, the attention maps also change to capture the dynamic action-specific relations. However, GA-SA weakens the physical and local interaction. A GCN branch is further introduced following MGCN~\cite{zou2021modulated}, whose parameters are updated during training and fixed during inference to capture the physical topology. 
The joint feature $X_{in} {\in} \mathbb{R}^{N \times d}$ enters each branch and is transformed by the attention maps and the adjacent matrices. By adding up the two features, the output contains the information from both static and dynamic graphs.

\subsection{Motion-Disentangled Regression Decoder}
\VspacePb

We design MDR including a Linear Blend Featuring~(LBF) module to learn vertex features through joint-vertex and vertex-vertex interactions, and a Motion-Disentangled Regression Head~(MDR Head) to predict base motions and use them to constitute vertex offsets. To avoid redundancies and make training more effcient~\cite{lin2021mesh, lin2021end}, MDR processes a coarse mesh with 431 vertices, then samples the vertex offsets up to 6K and adds to the original template mesh as the final result.

\noindent \textbf{Linear Blend Featuring.} Previous pose-based methods~\cite{choi2020pose2mesh, pqgcn, zheng2021lightweight} ignore the inherent joint-vertex relations in the transition from skeleton to mesh. In the algorithm of Linear Blend Skinning (LBS), each vertex is driven by joints and its coordinate can be represented as a weighted sum of all joints~\cite{jacobson2014skinning, loper2015smpl}.

Inspired by LBS, a cross-attention module is designed to perform joint-vertex interaction, which can be expressed as:
\vspace{-2mm}
\begin{equation}
    X^v_{cro} =\operatorname{softmax}({Q_v  K_j^{T}} / {\sqrt{d}})V_j, 
\end{equation}
\vspace{-5mm}
\begin{equation}
    Q_v=X^v_{in}W^c_Q, \quad K_j = X^j_{in}W^c_K, \quad V_j=X^j_{in}W^c_V,
    \vspace{-1mm}
\end{equation}
\sloppy
where $d$ is the feature dimension. $X^v_{in} {\in} \mathbb{R}^{V \times d}$ and $X^j_{in} {\in} \mathbb{R}^{N \times d} $ denote the input features of vertices and joints, $V$ and $N$ denote the numbers of vertices and joints, respectively. $W_Q^c, W_K^c, W_V^c {\in} \mathbb{R}^{d \times d}$ are the learnable weight matrices. Thus, each vertex feature is a weighted sum of the joint features. The input joint feature $X^j_{in}$ is the concatenation of 2D and 3D joint coordinates and the output joint feature from GAT. Besides, the input feature of a vertex $X^v_{in}$ is the concatenation of the vertex coordinate from the coarse template mesh and its nearest 3D joint coordinate.
After the joint-vertex interaction, we introduce a self-attention module for vertex-vertex interaction. 

\noindent \textbf{Motion-Disentangled Regression Head.} 
Traditional mesh regression heads project the high-dimension features to the vertex coordinates by a linear layer while ignoring the physical plausibility. 
Motivated by the observation in Sec.~\ref{sec:intro}, we propose a novel regression head based on the motion similarity in the same body part. Specifically, we predict several base motions to constitute the vertex offsets of coarse mesh (431 vertices) and add them to the original template mesh (6K vertices) after the upsampling operation. 
The coarse vertex offsets $\Delta M' {\in} \mathbb{R}^{V \times 3}$ consist of the weighted sum of base motions determining the general orientation and translation, and the learnable biases for refinement, which is expressed as:
\vspace{-2mm}
\begin{equation}
    \Delta M' ={\alpha} \times \operatorname{softmax}(M_A)M_B + M_C,
    \label{eq8}
    \vspace{-2mm}
\end{equation}
where $M_A {\in} \mathbb{R}^{V \times m}$ is the motion weight matrix, $M_B {\in} \mathbb{R}^{m \times 3}$ denotes $m$ base motions, $M_C {\in} \mathbb{R}^{V \times 3}$ denotes motion biases, and $\alpha {\in} \mathbb{R}^{V}$ denotes scaling factors. They are all learned from the network. The coarse vertex offset $\Delta M'$ is upsampled to the original resolution with 6K vertices through a linear projection and added to the template mesh to get the final mesh result.

\VspacePa
\subsection{Loss Functions} 
\VspacePb
GAT is first pretrained using the 3D joint loss to supervise the intermediate 3D pose.
Then following~\cite{choi2020pose2mesh, zheng2021lightweight}, the whole model is supervised by four losses: mesh vertex loss $\mathcal{L}_M$, 3D joint loss $\mathcal{L}_J$~(joints from the final mesh), surface normal loss $\mathcal{L}_N$, and surface edge loss $\mathcal{L}_E$. The total loss is calculated as:
\vspace{-2mm}
\begin{equation}
    \mathcal{L}=\lambda_M \mathcal{L}_M+\lambda_J \mathcal{L}_J+\lambda_N \mathcal{L}_N+\lambda_E \mathcal{L}_E,
    \vspace{-2mm}
\end{equation}
where $\lambda_{M} {=} 1$, $\lambda_{J} {=} 1$, $\lambda_{N} {=} 0.1$, $\lambda_{E} {=} 20$ in our experiments.

\begin{table}[t]
  \centering
  \setlength\tabcolsep{2pt} 
  \footnotesize
  \caption{Comparison with state-of-the-art methods on Human3.6M and 3DPW datasets. 
  $\S$ denotes the input is from 2D pose detectors~\cite{sun2018integral, zhang2020distribution}.
  * denotes the input is ground truth 2D pose. 
  None of these methods uses 3DPW for training.
  }
  \vspace{-3mm}
  \label{tab_3dpw}
  \scalebox{0.98}{
  \begin{tabular}{L{2mm}L{20mm}|C{9.5mm}C{14mm}|C{9.5mm}C{14mm}C{9mm}}
    \hline
     & \multirow{2}{*}{Method} & \multicolumn{2}{c|}{Human3.6M} & \multicolumn{3}{c}{3DPW} \\
     & & \scriptsize{MPJPE} \scriptsize{$\downarrow$} & \scriptsize{PA-MPJPE $\downarrow$} & \scriptsize{MPJPE $\downarrow$} & \scriptsize{PA-MPJPE $\downarrow$} & \scriptsize{MPVE $\downarrow$} \\
    \hline
    \multirow{7}{*}{\rotatebox{90}{\scriptsize{image}}} 
    & HMR~\cite{kanazawa2018end} & 88.0 & 56.8 & $-$ & 81.3 & 130.0 \\
    & GraphCMR~\cite{kolotouros2019convolutional} & $-$ &  50.1 & $-$ &  70.2 &  $-$ \\
    & SPIN~\cite{kolotouros2019learning} & 62.5 &  41.1  & 96.9 &  59.2 & 116.4  \\
    & PyMAF~\cite{zhang2021pymaf} & 57.7 & 40.5 & 92.8 & 58.9 & 110.1 \\
    & I2LMeshNet~\cite{moon2020i2l} & 55.7 & 41.1 &  93.2  &  57.7 & $-$  \\
    & ProHMR~\cite{kolotouros2021probabilistic} & $-$ & 41.2 & $-$ & 59.8 & $-$ \\
    & OCHMR~\cite{khirodkar2022occluded} & $-$ & $-$ & 89.7 & 58.3 & 107.1 \\    
    \hline
    \multirow{4}{*}{\rotatebox{90}{\scriptsize{video}}} 
    & VIBE~\cite{kocabas2020vibe} &  65.6  &  41.4  &  $93.5$ & $56.5$ & $113.4$ \\
    & TCMR~\cite{choi2021beyond} &  62.3  &  41.1  &  $95.0$ & $55.8$ & $111.5$ \\
    & AdvLearning~\cite{sun2022adversarial} &  $-$  &  $-$  &  92.6 & 55.2 & 111.9 \\
    & MPS-Net~\cite{wei2022capturing} &  $-$  & $-$ & 91.6 & 54.0 & 109.6 \\
    \hline
    \multirow{8}{*}{\rotatebox{90}{\scriptsize{2D pose}}} 
    & Pose2Mesh$\S$~\cite{choi2020pose2mesh} &  64.9  &  46.3 &  88.9  &  58.3 & 106.3  \\
    & PQ-GCN$\S$~\cite{pqgcn} &  64.6  &  47.9 &  89.2  &  58.3  & 106.4 \\
    & GTRS$\S$~\cite{zheng2021lightweight}  &  64.3  &  45.4 &  88.5  &  58.9 & 106.2\\
    & GATOR$\S$~(Ours) & 64.0 & 44.7 & 87.5 & 56.8 & 104.5 \\
    \cline{2-7}
    & Pose2Mesh*~\cite{choi2020pose2mesh}   &  51.3  &  35.9  &  65.1  &  34.6 &  $-$ \\
    & GTRS*~\cite{zheng2021lightweight} &  50.6  &  34.4 &  53.8  &  34.5 & 61.6 \\
    & GATOR*~(Ours) & \textbf{48.8} & \textbf{31.2} & \textbf{50.8} & \textbf{30.5} & \textbf{59.6}\\
    \hline
  \end{tabular}}
  \vspace{-3mm}
\end{table}

\vspace{-1mm}
\section{Experiments}\label{sec:experiments}
\vspace{-1mm}

\noindent \textbf{Datasets.} Human3.6M~\cite{ionescu2013human3}, 3DPW~\cite{von2018recovering}, COCO~\cite{lin2014microsoft}, and MuCo-3DHP~\cite{mehta2018single} are used following previous works~\cite{choi2020pose2mesh, pqgcn, zheng2021lightweight}. 

\noindent \textbf{Evaluation Metrics.} Three metrics are used to report the experimental results: Mean Per Joint Position Error~(MPJPE), Procrustes-Aligned MPJPE~(PA-MPJPE) that denotes MPJPE after rigid alinement, and Mean Per Vertex Error~(MPVE). 

\noindent \textbf{Implementation Details.} We first pretrain GAT and then train the whole GATOR in an end-to-end manner. The GAT is stacked by $L_1 {=} 6$ layers with the feature dimension $d_1 {=} 128$. In MDR, the LBF is stacked by $L_2 {=} 3$ layers with the feature dimension $d_2 {=} 64$. GAT is pretrained by Adam optimizer for 30 epochs with a batch size of 256 and a learning rate of $8 {\times} 10^{-4}$, while the whole GATOR is trained with a batch size of 64 and a learning rate of $1 {\times} 10^{-4}$ for 30 epochs. All experiments are conducted on one NVIDIA RTX 3090 GPU.

\subsection{Comparison with State-of-the-Art Methods}
\VspacePb
Table~\ref{tab_3dpw} compares GATOR with previous image/video/pose-based methods on Human3.6M and 3DPW datasets. For a fair comparison, we follow the same setting as previous works~\cite{zheng2021lightweight, choi2020pose2mesh, kanazawa2018end, kolotouros2019learning}. 
For Human3.6M, GATOR is trained on the Human3.6M training set and measured PA-MPJPE on the frontal camera set. 
For 3DPW, GATOR is trained on multi-datasets including Human3.6M, COCO, and MuCo-3DHP, and evaluated on the 3DPW testing set to examine cross-dataset generalization ability. 
When using detected 2D poses~\cite{sun2018integral, zhang2020distribution} as input, our method outperforms previous pose-based methods and achieves comparable results with image/video-based methods. 
Especially on 3DPW, although the input poses are not particularly accurate and without any image or temporal information, GATOR outperforms existing methods on the metrics of MPJPE and MPVE. 
When using GT 2D poses as inputs, the performance boosts by a large margin (improves 9.3\% from 34.4 mm to 31.2 mm on Human3.6M, 11.6\% from 34.5 mm to 30.5 mm on 3DPW in the metric of PA-MPJPE). It indicates that with a more accurate 2D pose detector, our method can further improve the performance.

\VspacePa
\subsection{Ablation Study}
\VspacePb

\noindent \textbf{Effectiveness of GA-SA and GCN.}
Table~\ref{tab-component} examines the components of GAT on the intermediate 3D pose and the final mesh.
The proposed GAT improves both pose and mesh performance, and a more accurate 3D pose is beneficial to higher mesh accuracy. 
Individual HE, PE or GCN brings similar improvements but combining them together boosts the performance by a clear margin (improve 2.3 mm in MPJPE and 3.2 mm in MPVE of 3D mesh). The improvements indicate that the hidden action-specific information explored by GA-SA and the physical skeleton topology extracted by GCN effectively complement each other to achieve better results.

\begin{table}[!t]
\begin{minipage}[t]{0.5\linewidth}
\footnotesize
  \setlength\tabcolsep{1pt}
  \centering
  \captionof{table}{Comparison of GAT components on 3DPW. }
  \vspace{-2mm}
  \scalebox{0.9}{
\begin{tabular}{C{3.5mm}C{3.5mm}C{6mm}|C{11mm}|C{9.5mm}C{9mm}}
    \hline
    \multirow{2}{*}{HE} & \multirow{2}{*}{PE} & \multirow{2}{*}{GCN} & 3D~Pose & \multicolumn{2}{c}{ 3D Mesh } \\
    & & & \scriptsize{MPJPE} $\downarrow$ & \scriptsize{MPJPE} $\downarrow$ & \scriptsize{MPVE} $\downarrow$ \\
    \hline
      &   &   & 87.1 & 89.8 & 107.7 \\
    \cmark &  &  & 86.5 &  88.6 & 105.5  \\
     & \cmark &  & 86.4 &  88.7 & 106.0  \\
      &  & \cmark & 86.6 & 88.7 & 105.6 \\
    \cmark & \cmark &  &  86.0  &  88.0 & 105.3 \\
    \cmark &  & \cmark  & 85.9  &  88.3 &  105.3 \\
     & \cmark & \cmark  & 85.2  &  88.0 &  105.1 \\
    \cmark & \cmark & \cmark  & \textbf{84.3}  &  \textbf{87.5} &  \textbf{104.5} \\
    \hline
    \VspaceTc
  \end{tabular}}
  \label{tab-component}
\end{minipage}
\hfill
\begin{minipage}[t]{0.45\linewidth}
  \footnotesize
  \setlength\tabcolsep{1pt}
  \centering
  
  \captionof{table}{Comparison of different regressors on 3DPW.}
  \vspace{-2mm}
  \scalebox{0.9}{
\begin{tabular}{l|C{14mm}C{9mm}}
    \hline
    \multirow{2}{*}{Regressor} & \multicolumn{2}{c}{ 3D Mesh } \\
    &   \scriptsize{PA-MPJPE$\downarrow$} & \scriptsize{MPVE} $\downarrow$ \\
    \hline
    Linear 
    & 63.2 & 119.9 \\
    Linear + LBF
    & 58.8 & 107.7 \\
    MDR$^{20}$ w/o LBF
    & 58.2 & 107.2  \\ 
    MDR$^5$
    & 56.8 & 104.8 \\
    MDR$^{10}$
    & 57.4 & 105.5 \\
    MDR$^{20}$
    &  \textbf{56.8} &  \textbf{104.5} \\
    MDR$^{30}$
    & 57.4 & 105.2 \\
    MDR$^{40}$
    & 57.5 & 105.5 \\
    \hline
  \end{tabular}}
  \label{tab-regressor}
\end{minipage}
\end{table}

\begin{figure}[!t]
  \vspace{-2mm}
  \includegraphics[width=\linewidth]{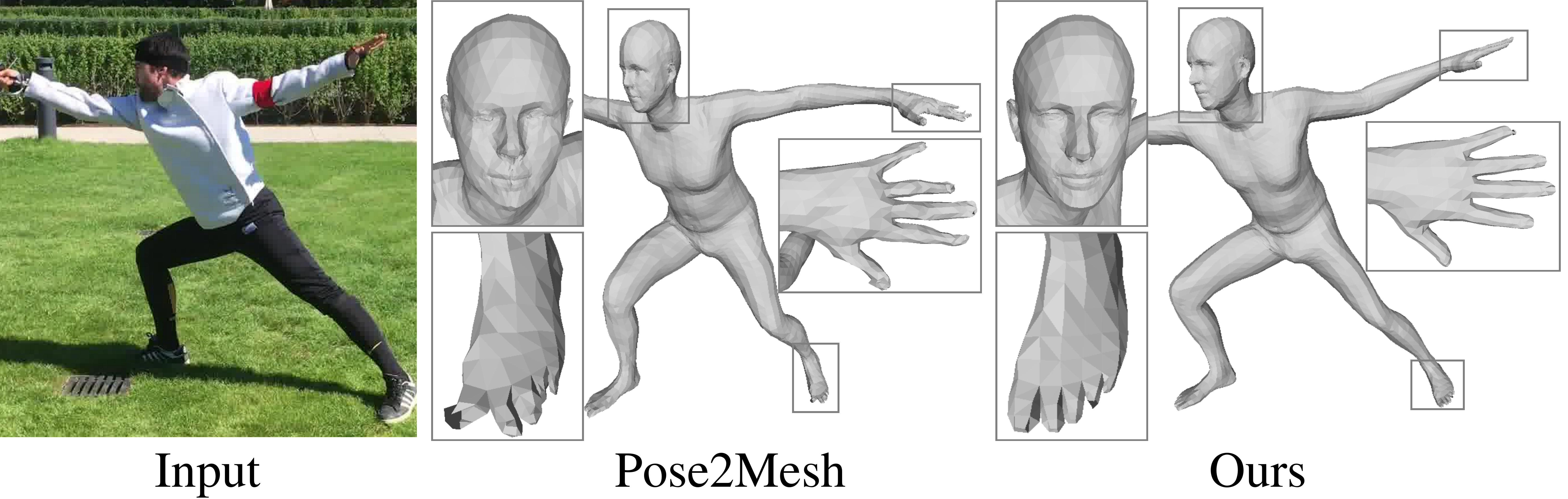}
  \vspace{-5mm}
  \caption{Qualitative comparison of mesh details between Pose2Mesh~\cite{choi2020pose2mesh} and our proposed GATOR.}
  \label{fig_mesh}
  \vspace{-4mm}
\end{figure}

\noindent \textbf{Effectiveness of MDR.}
Table~\ref{tab-regressor} evaluates the impact of MDR by removing LBF, replacing MDR Head with a general linear layer, and setting different base motion numbers. The top two lines show that the joint and vertex interactions in LBF are effective for relations exploring and significantly improve the results. When considering MDR Heads, the performance is further improved, where MDR Head with 20 base motions achieves the optimal results. 
More or fewer base motions may bring inappropriate clustering, which drops the performance.

\VspacePa
\subsection{Qualitative Results}
\VspacePb
Fig.~\ref{fig_vis} shows qualitative results compared to GraphCMR~\cite{kolotouros2019convolutional}, I2L-MeshNet~\cite{moon2020i2l}, and GTRS~\cite{zheng2021lightweight} on the COCO dataset. The first two are image-based methods that are often impacted by the background, while the latter two pose-based methods are more robust, whereas GATOR produces more plausible results.
Fig.~\ref{fig_mesh} compares mesh details between Pose2Mesh~\cite{choi2020pose2mesh} and GATOR. Pose2Mesh is prone to generate artifacts due to the sub-optimal prediction of vertex positions, while our method can provide more accurate vertices for fine-grained meshes.

\vspace{-2mm}
\section{Conclusion}
\vspace{-2mm}
In this paper, we present GATOR, a novel network for 3D human mesh recovery from a 2D pose. To explore multiple joint and vertex relations in the evolution from skeleton to mesh, GAT is designed to explore joint relations by combining a GCN branch and a GA-SA branch in parallel for static and dynamic graph learning, where GA-SA takes two important skeleton encodings to enhance the graph awareness. Besides, MDR is proposed to model joint-vertex and vertex-vertex interactions and generate the vertex coordinates in a motion-disentangled regression, 
which provides more accurate results.
Extensive experiments show that our method achieves state-of-the-art performance on two challenging benchmarks.

\vfill\pagebreak

{\small
\bibliographystyle{IEEEbib}
\bibliography{ref}

}

\end{document}